# A Multi-modal Early Detection System for Infrastructure-based Freight Signal Priority

Ziyan Zhang[1]; Chuheng Wei[2]; Xuanpeng Zhao[3]; Siyan Li[4]; Will Snyder[5]; Mike Stas[6]; Peng Hao, Ph.D.[7]; Kanok Boriboonsomsin, Ph.D.[8]; and Guoyuan Wu, Ph.D.[9]

[1]Ph.D. Student, Center for Environmental Research and Technology (CE-CERT), University of California at Riverside (UCR), Riverside, CA (corresponding author). ORCID: https://orcid.org/0000-0001-7977-4110. Email: zzhan554@ucr.edu
[2]Ph.D. Candidate, CE-CERT, UCR, Riverside, CA. Email: chuheng.wei@email.ucr.edu
[3]Ph.D., CE-CERT, UCR, Riverside, CA. Email: xzhao094@ucr.edu
[4]Ph.D. Student, CE-CERT, UCR, Riverside, CA. Email: sli442@ucr.edu
[5]Ph.D. Student, CE-CERT, UCR, Riverside, CA. Email: rsnyd009@ucr.edu
[6]Ph.D. Candidate, CE-CERT, UCR, Riverside, CA. Email: mstas001@ucr.edu
[7]Associate Research Engineer, CE-CERT, UCR, Riverside, CA. Email: peng.hao@ucr.edu
[8]Research Engineer, CE-CERT, UCR, Riverside, CA. Email: kanok.boriboonsomsin@ucr.edu
[9]Researcher, CE-CERT, UCR, Riverside, CA. Email: guoyuan.wu@ucr.edu

**ABSTRACT**

Freight vehicles approaching signalized intersections require reliable detection and motion estimation to support infrastructure-based Freight Signal Priority (FSP). Accurate and timely perception of vehicle type, position, and speed is essential for enabling effective priority control strategies. This paper presents the design, deployment, and evaluation of an infrastructure-based multi-modal freight vehicle detection system integrating LiDAR and camera sensors. A hybrid sensing architecture is adopted, consisting of an intersection-mounted subsystem and a midblock subsystem, connected via wireless communication for synchronized data transmission. The perception pipeline incorporates both clustering-based and deep learning-based detection methods with Kalman filter tracking to achieve stable real-time performance. LiDAR measurements are registered into geodetic reference frames to support lane-level localization and consistent vehicle tracking. Field evaluations demonstrate that the system can reliably monitor freight vehicle movements at high spatio-temporal resolution. The design and deployment provide practical insights for developing infrastructure-based sensing systems to support FSP applications.

**INTRODUCTION**

Roadside sensing using Light Detection And Ranging (LiDAR) and cameras has gained increasing attention for transportation applications such as traffic monitoring and cooperative perception. In this work, we focus on freight signal priority (FSP) and leverage infrastructure-based sensing to estimate vehicle time of arrival (ToA) for priority requests.

FSP aims to improve truck mobility at signalized intersections while accounting for their slow acceleration and high fuel cost in stop-and-go conditions. Despite accounting for only ~9% of vehicles and ~17% of VMT, heavy-duty trucks contribute ~39% of life-cycle GHG emissions (Moultak et al., 2017). Eco-FSP approaches have shown 5–10% network-wide fuel savings and up to 26% travel time reduction, with connected-truck systems reporting up to 25.3% fuel and $CO_2$ reduction (Kari et al., 2014; Park et al., 2019). However, deployments relying on V2I/V2V communication require onboard equipment, limiting real-world adoption.

To overcome these challenges, we develop an infrastructure-based, multi-modal freight vehicle detection system integrating LiDAR and cameras. The system contains two subsystems tailored to site-specific terrain constraints: (1) a long-range LiDAR subsystem with camera installed at the intersection, and (2) a mid-range LiDAR subsystem with camera placed midblock to overcome visibility blockage caused by roadway curvature. Wireless communication transmits midblock detection results to the intersection for downstream FSP operations.

The key contributions of this work are summarized as follows:
- A complete design, implementation, and evaluation pipeline for an infrastructure-based, multi-modal freight vehicle detection system is presented.
- A dual-subsystem detection architecture is proposed to address the unique challenges of the FSP deployment site, particularly limited visibility caused by roadway curvature.
- Data acquisition and processing pipelines are developed, and preliminary detection and tracking results are demonstrated and evaluated.
- LiDAR-derived coordinates is registered with geodetic reference frames, enabling lane-level vehicle localization.

**SYSTEM DESIGN**

As shown in Figure 1, the FSP system is deployed at the main intersection with pole-mounted sensors providing long-range coverage in two opposing directions. However, roadway curvature in one direction causes partial occlusion, limiting far-distance visibility. To mitigate this, a mid-range detection subsystem is installed at the midblock using the existing pole and cabinet to detect freight vehicles earlier. This two-site setup introduces an additional requirement for reliable wireless communication between the midblock and the intersection. Specifically, Figure 1(c) presents a bird's-eye view illustrating the terrain and subsystem placement.

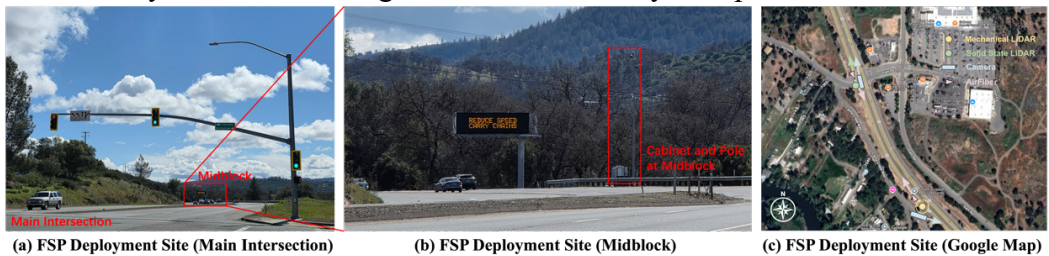

(a) FSP Deployment Site (Main Intersection)　(b) FSP Deployment Site (Midblock)　(c) FSP Deployment Site (Google Map)

**Figure 1. An Overview in the FSP Deployment Site.**

The system consists of two subsystems—one at the main intersection and the other at the midblock—as shown in Figure 2. A LiDAR–camera pair with a long-range solid-state LiDAR (Livox TELE-15) is installed at the intersection to monitor the southbound direction. To address the occlusion caused by horizontal roadway curvatures, a second LiDAR–camera pair with a mid-range mechanical LiDAR (Robosense Ruby Plus) is deployed at the midblock to ensure continuous detection of approaching traffic. A wireless communication bridge (Ubiquiti AirFiber 5XHD) enables data transmission between the two subsystems.

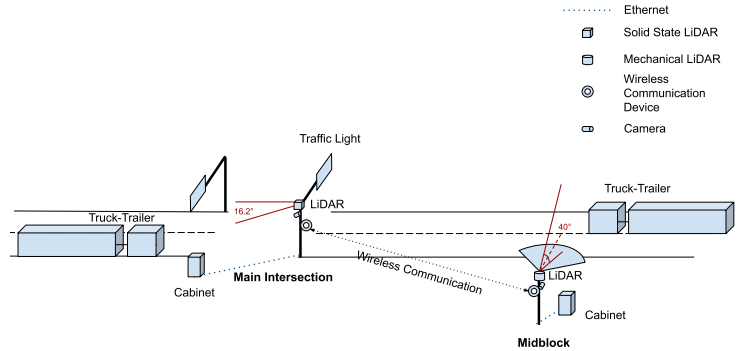

**Figure 2. The Devices Mounted at the Main Intersection and the Midblock.**

For the detection system components, a mid-range sensing unit is deployed at the midblock to monitor vehicles approaching from the south. It includes detection sensors, a PoE switch, an edge computer, and a point-to-point wireless transmitter. The edge computer processes LiDAR point clouds locally to obtain vehicle detection results, which are then transmitted to the main intersection via a Precision Time Protocol (PtP) wireless bridge, with time synchronization maintained using PTP (Cho et al., 2009). At the main intersection, the edge computer receives detections from both directions and generates freight priority requests for the traffic controller.

## DATA ACQUISITION AND PROCESSING

### Raw Data Acquisition

*Detection Data from Sensors*

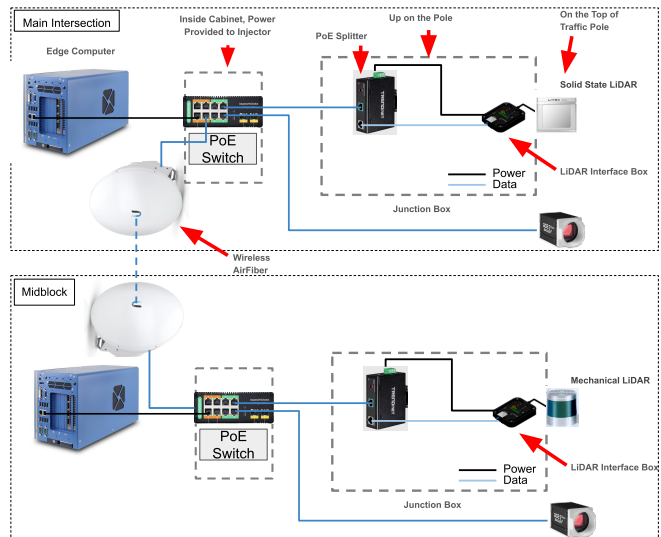

**Figure 3. Real Integration Details of the FSP Detection System.**

As shown in Figure 3, the LiDAR, camera, and AirFiber are connected and powered through a Power-over-Ethernet (PoE) switch. Because the LiDAR does not support native PoE, a PoE splitter is used to merge power and data into a single cable. The camera images and LiDAR point clouds are transmitted through the PoE switch to the edge computer for data acquisition. The AirFiber link is primarily used to send detection results and will not be further discussed in this paper.

Camera and LiDAR data are collected using ROS2 with timestamps synchronized to the system clock. Figure 4 shows example data captured at the main intersection and the midblock, where freight vehicles are clearly visible in both the images and the point clouds.

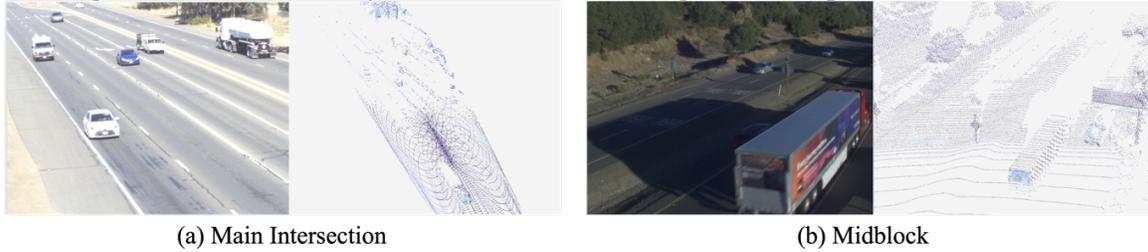

(a) Main Intersection  (b) Midblock

**Figure 4. Example of Data Collected at the Main Intersection and the Midblock.**

*GPS Calibration Data*

For GPS calibration, we register the LiDAR point cloud to global coordinates (latitude/longitude). An OXTS AV200 GNSS unit is mounted on a Toyota Corolla, and GPS data is recorded alongside LiDAR and camera data. The LiDAR is used for geolocation alignment, while camera images help identify the vehicle in the point cloud. Both static reference points and continuous trajectory data are collected for calibration, as shown in Figure 5.

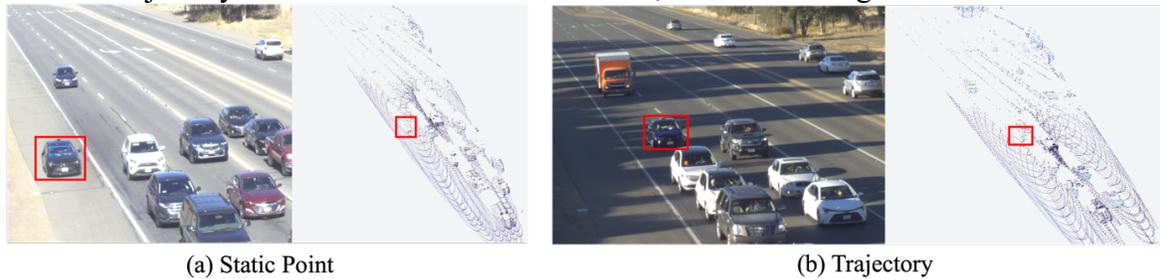

(a) Static Point  (b) Trajectory

**Figure 5. Recorded Static and Trajectory Data (GPS and Point Cloud).**

**Data Processing**

*Detection and Tracking*

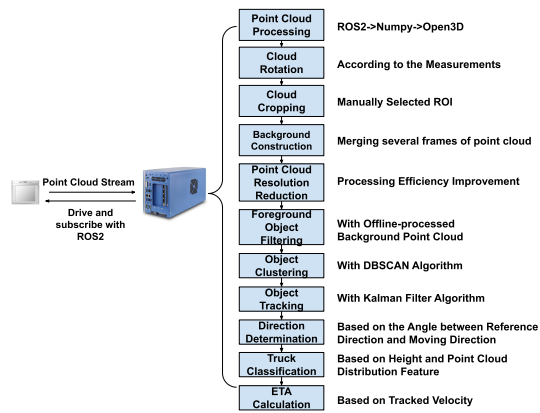

**Figure 6. Conventional Detection Pipeline.**

With the field data collected, we evaluate off-the-shelf detection and tracking algorithms. A DL-based detector and a conventional LiDAR-only method are considered; however, pretrained models such as PointPillars (Lang et al., 2019) trained on NuScenes (Caesar et al., 2020) failed to generalize to the collected data, especially for solid-state LiDAR due to different scanning patterns. Therefore, we adopt a conventional LiDAR-based detector with a simple tracking algorithm in this study.

The pipeline is shown in Figure 6. Raw point clouds are ground-aligned, cropped by ROI, denoised, and downsampled for efficiency. Foreground points are extracted by subtracting an offline background map, and clusters are generated via DBSCAN (Ester et al., 1996).

Detection and classification leverage both per-frame results and multi-frame tracking. Tracking across frames enables motion direction estimation, truck identification, and ToA computation based on tracked velocity.

**Cloud Rotation**: The roadside LiDAR is installed with non-zero roll and pitch, causing the raw point cloud to appear tilted. To make the z-axis vertical and enable downstream processing, we apply a rotation correction. Let $\phi$ and $\theta$ denote roll and pitch. The rotation matrices are:

$$R_x(\phi) = \begin{pmatrix} 1 & 0 & 0 \\ 0 & \cos\phi & -\sin\phi \\ 0 & \sin\phi & \cos\phi \end{pmatrix}, \quad R_y(\theta) = \begin{pmatrix} \cos\theta & 0 & \sin\theta \\ 0 & 1 & 0 \\ -\sin\theta & 0 & \cos\theta \end{pmatrix}. \tag{1}$$

Pitch is typically dominant for roadside mounting, so we apply correction in Pitch → Roll order:

$$R = R_y(\theta)\, R_x(\phi) \tag{2}$$

Each LiDAR point p = [x,y,z]$^T$ is corrected by:

$$p' = R^T p \tag{3}$$

where p′ is the corrected point cloud coordinate. yielding a leveled point cloud with an upright z-axis. This improves height consistency and enables more reliable truck–car differentiation based on vertical geometry (see Figure 7).

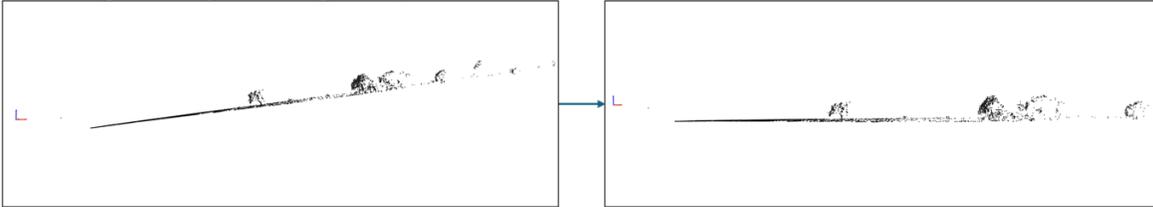

**Figure 7. Cloud Rotation (Solid State LiDAR)**

**Cloud Cropping**: In Figure 8, the point cloud is cropped using a manually defined ROI to remove background clutter and reduce processing load.

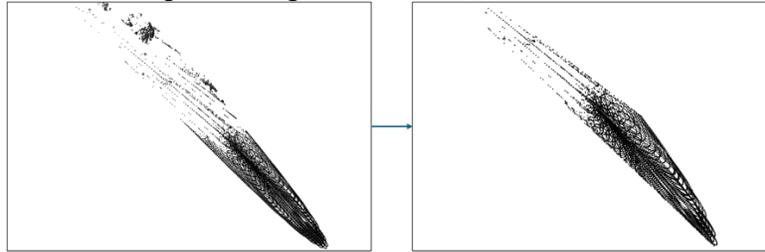

**Figure 8. Cloud Cropping (Solid State LiDAR).**

**Background Construction**: Due to the scanning pattern of the solid-state LiDAR, point distributions vary across frames, causing outliers when using a single offline background cloud for foreground extraction. To address this, we merge multiple background frames to build a more complete background map.

**Point Cloud Resolution Reduction**: Given the high LiDAR resolution (~14k points per frame), voxel-grid downsampling is applied to the raw point cloud $\mathcal{P}$ for real-time processing. The 3D space is divided into cubic voxels of size $s_v$ and each point $p_i = (x_i, y_i, z_i)$ is mapped to a voxel index $k_i = \left(\left\lfloor\frac{x_i}{s_v}\right\rfloor, \left\lfloor\frac{y_i}{s_v}\right\rfloor, \left\lfloor\frac{z_i}{s_v}\right\rfloor\right)$. For each occupied voxel k, we compute the centroid $c_k = \frac{1}{|\mathcal{P}_k|}\sum_{p_i \in \mathcal{P}_k} p_i$ and the downsampled point cloud is given by $\mathcal{P}' = \{c_k \mid \mathcal{P}_k \neq \emptyset\}$, effectively reducing the point cloud resolution while preserving the overall geometric structure.

**Foreground Object Filtering**: To extract moving objects from the downsampled point cloud $\mathcal{P}'$, we perform background distance filtering using a KD-tree (Bentley 1975). For each point $p_i \in \mathcal{P}'$, we compute the nearest background distance $d_i = \min_{b \in \mathcal{B}}|\boldsymbol{p_i} - \boldsymbol{b}|_2$ with a background map $\mathcal{B}$. Let $\mu_d$ and $\sigma_d$ denote the mean and standard deviation of $\{d_i\}$. Points whose distance exceeds an adaptive threshold $\tau = \mu_d + \alpha\sigma_d$ (with $\alpha > 0$ controlling sensitivity) are classified as foreground. The resulting foreground set is $\mathcal{F} = \{p_i \mid d_i > \tau\}$, which effectively removes background structures (road surfaces, vegetations, and static infrastructure) while retaining vehicles and other dynamic objects.

**Object Clustering**: After foreground extraction, object instances are grouped using the DBSCAN (Ester et al. 1996) clustering method. For each point $\boldsymbol{p_i}$, the ε-neighbourhood is defined as $\mathcal{N}_\varepsilon(p_i) = \{p_j \in P \mid ||p_j - p_i|| \leq \varepsilon\}$. A point is considered as a core point if the number of points within a radius ε exceeds a minimum density threshold (MinPts), as $|\mathcal{N}_\varepsilon(\boldsymbol{p_i})| \geq \{MinPts\}$. Points that are directly or indirectly density-connected to core points are assigned to the same cluster $\mathcal{C} = \{C_k \mid k = 1, \ldots, K\}$, where $C_k = \{\boldsymbol{p_i} \in P \mid \boldsymbol{p_i} \text{ belongs to cluster } k\}$, while isolated points are treated as noise. For each resulting cluster C_k, we compute its centroid and height statistics as,

$$c_k = \frac{1}{|C_k|}\sum_{p_i \in C_k} bp_i \quad (4)$$

$$h_k^{max} = \max_{p_i \in C_k} z_i \quad (5)$$

$$\sigma_{z,k} = \text{std}(\{z_i \mid p_i \in C_k\}) \quad (6)$$

where includes maximum point height and vertical variation. These features are later used for vehicle classification and tracking.

**Object Tracking**: For object tracking, we adopt a 6-dimensional constant-velocity Kalman Filter (Welch 1995) to smooth trajectories and handle short occlusions.. The state vector is defined as $\mathbf{x} = [x, y, z, v_x, v_y, v_z]^T$, and only the 3D position is observable through LiDAR measurements via $\mathbf{z} = [x, y, z]$. The state evolves following a linear motion model with sampling time $\Delta t$:

$$\mathbf{x}_k = \mathbf{A}\mathbf{x}_{k-1} + \mathbf{w}_k, \quad \mathbf{z}_k = \mathbf{H}\mathbf{x}_k + \mathbf{v}_k \quad (7)$$

where

$$\mathbf{A} = \begin{bmatrix} 1 & 0 & 0 & \Delta t & 0 & 0 \\ 0 & 1 & 0 & 0 & \Delta t & 0 \\ 0 & 0 & 1 & 0 & 0 & \Delta t \\ 0 & 0 & 0 & 1 & 0 & 0 \\ 0 & 0 & 0 & 0 & 1 & 0 \\ 0 & 0 & 0 & 0 & 0 & 1 \end{bmatrix}, \mathbf{H} = \begin{bmatrix} 1 & 0 & 0 & 0 & 0 & 0 \\ 0 & 1 & 0 & 0 & 0 & 0 \\ 0 & 0 & 1 & 0 & 0 & 0 \end{bmatrix};$$

$\mathbf{w}_k$ and $\mathbf{v}_k$ denote the process noise (object acceleration and motion uncertainty) and measurement noise introduced by LiDAR measurement, with covariances $\mathbf{Q} = \text{cov}(\mathbf{w}_k)$ and $\mathbf{R} = \text{cov}(\mathbf{v}_k)$.

At each frame, the filter first performs prediction using the motion model:

$$\hat{\mathbf{x}}_{k|k-1} = \mathbf{A}\hat{\mathbf{x}}_{k-1|k-1}, \quad \mathbf{P}_{k|k-1} = \mathbf{A}\mathbf{P}_{k-1|k-1}\mathbf{A}^T + \mathbf{Q} \quad (8)$$

where $\mathbf{P}$ is the state estimation error covariance, enlarged during prediction by $\mathbf{Q}$.

Then the filter updates the state using the new measurement:

$$\hat{x}_{k|k} = \hat{x}_{k|k-1} + K_k(z_k - H\hat{x}_{k|k-1}), P_{k|k} = (I - K_kH)P_{k|k-1} \quad (9)$$

where $K_k$ denotes the Kalman gain:

$$K_k = P_{k|k-1}H^T(HP_{k|k-1}H^T + R)^{-1} \quad (10)$$

This filter smooths trajectories, reduces jitter caused by sparse point clusters, and provides stable velocity estimation for downstream classification.

**Direction Determination**: As shown in Figure 9, the relative motion direction is inferred from the dot product between the object's velocity vector *a* and the line-of-sight vector *b*. Specifically,

$$a \cdot b \begin{cases} > 0, & object\ approaching \\ < 0, & object\ departing \end{cases} \quad (11)$$

This sign-based criterion provides a compact and mathematically robust formulation for direction classification.

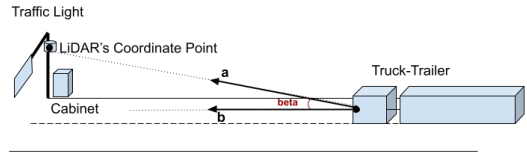

**Figure 9. Direction Determination Schematic.**

**Truck Classification**: For truck classification, we select three key features: absolute height, relative height, and the Z-axis point distribution. As shown in Figure 10, the relative height corresponds to the object's Z-coordinate in the LiDAR coordinate frame. The absolute height is defined as the maximum vertical difference among all points within a cluster. In Figure 10, the absolute height, relative height, and Z-axis point distribution of Cluster 4 are noticeably different from those of the other clusters, indicating that this cluster corresponds to a truck in the real world. In comparison, Clusters 0–3 represent passenger vehicles.

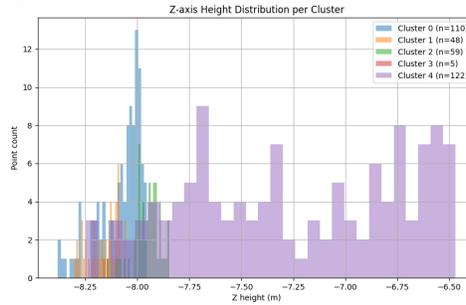

**Figure 10. Truck Classification Feature Distribution.**

In Figure 11, a foreground object is clustered and tracked over 10 frames using DBSCAN and the Kalman Filter, producing a smooth trajectory of an approaching vehicle represented by a green line.

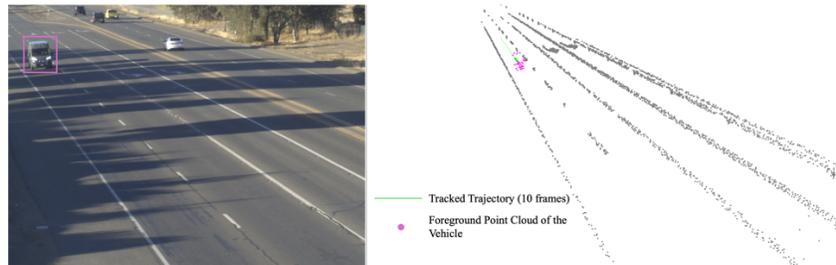

**Figure 11. A Showcase Example of the Detection and Tracking Algorithm.**

*GPS Calibration*

The pipeline is shown in Figure 12. To obtain an initial extrinsic transformation between the LiDAR frame L and the global ENU frame E, several stationary segments are extracted from the GPS and LiDAR logs. GPS positions are converted to ENU, and corresponding static targets in the LiDAR cloud are manually labeled. Given $N$ paired samples $\{(\mathbf{p}_{\text{LiDAR}}(i), \mathbf{p}_{\text{ENU}}(i))\}_{i=1}^{N}$, the initial rigid transformation is estimated via point-to-point registration:

$$\mathbf{p}_{\text{ENU}} = R_0\, \mathbf{p}_{\text{LiDAR}} + t_0 \tag{12}$$

$$(\mathbf{R}_0, \mathbf{t}_0) = \arg\min_{\mathbf{R},\mathbf{t}} \sum_{i=1}^{N} \|\mathbf{R}\mathbf{p}_{\text{LiDAR}}(i) + \mathbf{t} - \mathbf{p}_{\text{ENU}}(i)\|^2,\ \mathbf{R} \in SO(3) \tag{13}$$

The resulting $(\mathbf{R}_0, \mathbf{t}_0)$ maps LiDAR points into ENU coordinates.

Using this initial transform, a moving vehicle is tracked in LiDAR and converted into ENU, while the GPS log provides a reference trajectory. Both trajectories are resampled by arc-length and aligned without requiring time synchronization. A 2D rigid alignment refines yaw and planar translation:

$$\mathbf{p}'_{\text{ENU}} = \begin{bmatrix} \mathbf{R}_z & \mathbf{t}_{xy} \\ \mathbf{0}^\top & 1 \end{bmatrix} \mathbf{p}_{\text{ENU}},\ \mathbf{R}_z = \begin{bmatrix} \cos\theta & -\sin\theta \\ \sin\theta & \cos\theta \end{bmatrix} \tag{14}$$

A vertical offset restores height consistency (considering roof-vs-centroid height difference):

$$\mathbf{T}_{\text{ref}} = \mathbf{t}_{xy} + [0,0,\Delta z]^\top,\ \Delta z = \bar{z}_{\text{GPS}} - \bar{z}_{\text{LiDAR}} \tag{15}$$

The final refined extrinsic is:

$$\mathbf{R}_{\text{final}} = \mathbf{R}_z \mathbf{R}_0,\ \mathbf{t}_{\text{final}} = \mathbf{R}_z \mathbf{t}_0 + \mathbf{T}_{\text{ref}} \tag{16}$$

This two-stage procedure combines static registration and dynamic trajectory alignment, producing a robust GPS–LiDAR extrinsic.

Figure 13 shows an example overlay of the transformed GPS trajectory (red dotted line) on the LiDAR point cloud.

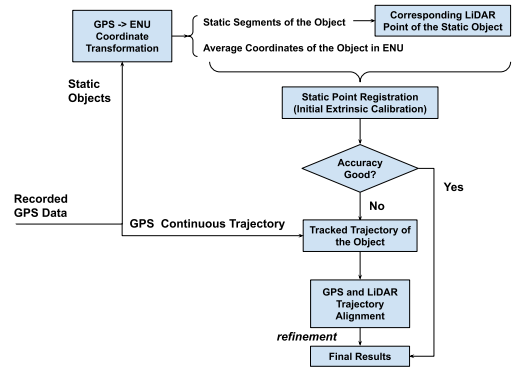

Figure 12. Diagram of GPS Calibration Method.

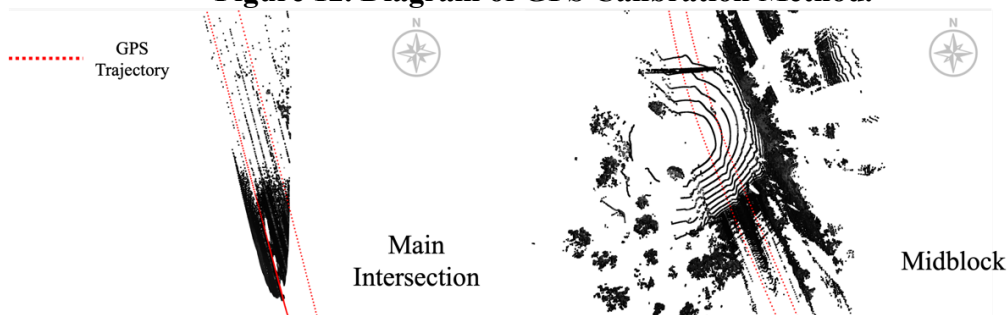

Figure 13. GPS Calibration Results (Site 6, Main Intersection).

## EVALUATION

The system is evaluated in terms of detection accuracy, real-time efficiency, and GPS calibration accuracy.

**Detection Accuracy for FSP**

We selected 20 representative detection scenarios, including trucks (e.g., standard and concrete trucks) and non-truck vehicles that can be misidentified as trucks (e.g., vans, pickup trucks with trailers, bucket trucks). One representative frame is chosen per scenario, without bias toward favorable cases, covering varied viewing angles and distances for a comprehensive evaluation.

*Ground-Truth Annotation*

Each frame contains one primary target, which we manually annotate as either a truck (Long or Compact) to evaluate correct FSP triggering, or a non-truck to assess false triggers. Other vehicles are ignored since full-scene labeling is unnecessary for FSP evaluation.

*Detection Output*

The conventional LiDAR-based detector outputs object-level classification results and 3D point cluster centers without bounding boxes. To simplify evaluation and to align with the traffic priority application context, we only focus on whether the system correctly recognizes a truck within the frame, regardless of other detections.

*Evaluation Logic*

For each scenario frame, we determine whether truck priority should be triggered based on frame-level outcomes and formulate a frame-based confusion matrix in Table 1.

**Table 1. Confusion Matrix.**

| GT / Prediction | Predicted-Truck Exists | No Predicted-Truck |
|---|---|---|
| GT: Truck | TP (matched within threshold) | FN (no matched prediction) |
| GT: Non-truck | FP (any predicted truck) | TN (none predicted) |

Here, Long Truck and Compact Truck are treated separately due to size differences, using different spatial thresholds: Long Truck Threshold = 10m; Compact Truck Threshold = 4m.

This design compensates for partial visibility when observing large trailers from one side, where the cluster-derived center often deviates significantly from the vehicle's true geometric center.

Results of 20 scenarios: TP=6, FP=2, FN=6, TN=6.

*Metric Computation*

Using the above definitions, we compute standard binary classification metrics to evaluate correct truck-triggering and suppression of false activations.

$$\text{Precision} = \frac{TP}{TP + FP} \tag{17}$$

$$\text{Recall} = \frac{TP}{TP + FN} \tag{18}$$

$$F_1 = \frac{2 \cdot \text{Precision} \cdot \text{Recall}}{\text{Precision} + \text{Recall}} \tag{19}$$

**Table 2. Evaluation Results.**

| Precision | Recall | F1 Score |
|---|---|---|
| 0.75 | 0.50 | 0.60 |

Table 2 shows the evaluation results with high precision but lower recall, meaning the system conservatively triggers FSP yet misses some trucks. Future work can focus on improving direction-aware matching and track-level classification to further reduce false negatives, especially for long vehicles.

**Real-time Efficiency Evaluation**

Besides stable detection accuracy, real-time performance is essential for field deployment. We evaluated the relationship between foreground point count and per-frame processing time to meet the 10 Hz requirement. Approximately 5 minutes of data at the main intersection and 3 minutes at the midblock are replayed. As shown in Figure 14 and Figure 15, processing time increases with foreground density but remains below 0.05 s per frame, well within real-time constraints.

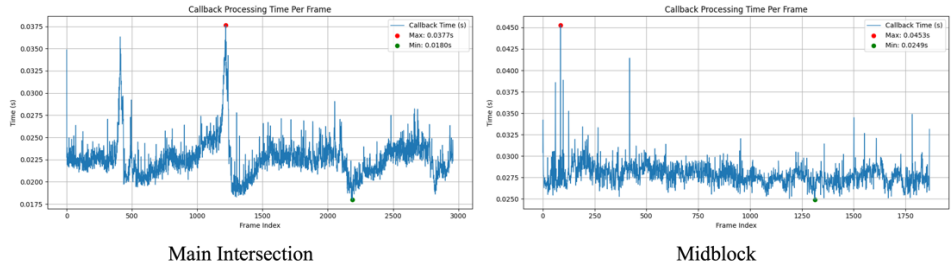

**Figure 14. Processing Time Per Frame.**

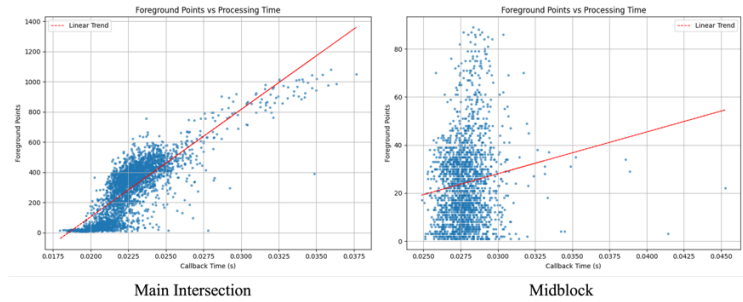

**Figure 15. Correlation between Number of Foreground Points and Processing Time.**

**GPS Calibration Evaluation**

GPS Calibration quality is measured using static registration and trajectory alignment error as shown in the following sections.

*Static Point Registration Error*

For each manually selected static point pair, we compute the static registration error $e_i^{\text{static}}$, and summarize the mean and maximum values $\bar{e}^{\text{static}}$ and $e_{max}^{\text{static}}$.

$$e_i^{\text{static}} = \|\mathbf{R}\mathbf{p}_i^L + \mathbf{T} - \mathbf{p}_i^E\|_2 \tag{20}$$

*Dynamic Trajectory Alignment Error*

After trajectory-based refinement, we compare the LiDAR and GPS trajectories in ENU using arc-length resampling. The horizontal trajectory error $e_i^{\text{traj}}$ and report its mean and maximum values $\bar{e}^{\text{traj}}$ and $e_{max}^{\text{traj}}$. Table 3 presents the static point registration errors and dynamic trajectory alignment errors for the two sites.

$$e_i^{\text{traj}} = |\tilde{q}_{i,xy}^E - q_{i,xy}^E|_2 = \sqrt{(\tilde{x}_i - x_i)^2 + (\tilde{y}_i - y_i)^2} \tag{21}$$

**Table 3. GPS Calibration Evaluation Results**

| Site | Point Registration Error (m) | | Trajectory Alignment Error (m) | |
|---|---|---|---|---|
| | Mean | Max | Mean | Max |
| Main Intersection (6 points, 1 trajectory) | 0.210 | 0.342 | 0.345 | 0.802 |
| Midblock (6 points, 2 trajectories) | 0.346 | 0.746 | 0.885 | 3.356 |

The midblock calibration shows noticeably higher registration and trajectory alignment errors compared to the main intersection. This is mainly because the static reference object is located farther from the LiDAR, resulting in a sparser point cloud and weaker geometric constraints during static point registration. Although the midblock trajectories have similar total length, vehicles travel faster in this section, leading to fewer sampled points along the trajectory. With larger spacing between sequential points, the curvature becomes less detailed, reducing the robustness of trajectory-based refinement. Moreover, alignment of two separated trajectories may accumulate bias, especially when the initial extrinsic contains slight yaw or scale drift, further amplifying errors in the midblock scenario.

**DISCUSSION**

With the system design and data processing pipeline established, several aspects merit further discussion.

**System Design**: The two detection subsystems are designed for a curved site requiring early truck detection. However, each module can operate independently or be combined differently for deployment at other locations.

**Algorithms**: Due to limited labeled data and unsatisfactory performance of pretrained models, deep learning methods are not fully explored. Future work will construct a multi-modal dataset (LiDAR + camera) with long-range freight detection and solid-state LiDAR characteristics—features rarely available in existing vehicle datasets, especially for roadside sensing. Camera information remains underutilized; daytime camera–LiDAR fusion could enhance detection accuracy, particularly for DL-based approaches (Wei et al., 2025).

**GPS Calibration**: Most static points used for calibration are collected along a near-collinear path, reducing geometric constraint strength and degrading initial transformation accuracy. Collecting static references across wider spatial distributions, ideally covering a plane, would improve calibration robustness.

**FSP-based Evaluation**: Frame-level evaluation directly reflects the practical objective of FSP—whether a truck should trigger priority—without requiring full bounding box annotation. This metric aligns well with real deployment needs and avoids unnecessary labeling overhead.

**CONCLUSION**

In this paper, we presented an infrastructure-based early detection system for freight signal priority, including system design, data acquisition, LiDAR–camera processing, detection/tracking, and GPS–LiDAR calibration. Visualization and real-time tests verified system feasibility and efficiency, while calibration achieved lane-level localization accuracy. Future work will focus on improving detection through multi-modal fusion and developing a well-annotated dataset to support deep learning–based models.